# A KERNEL FISHER DISCRIMINANT ANALYSIS-BASED TREE ENSEMBLE CLASSIFIER: KFDA FOREST

**Donghwan Kim[1], Seung Hwan Park[2], and Jun-Geol Baek[1*]**

[1] School of Industrial Management Engineering, Korea University
Seoul, Korea

[2] School of Mechanical Engineering, Chungnam National University
Daejeon, Korea
[*]Corresponding author's e-mail: jungeol@korea.ac.kr

In general, an ensemble classifier is more accurate than a single classifier. In this study, we propose an ensemble classifier called the kernel Fisher discriminant analysis forest (KFDA Forest), which is a tree-based ensemble method that applies KFDA. To promote diversity, bootstrap is used, and variable sets are randomly divided into $K$ subsets. KFDA is performed on each subset to increase classification accuracy. KFDA maximizes the distance between classes while minimizing the distance within classes. KFDA can also be applied to classification problems in a nonlinear data structure using the kernel trick because it can transform the input space into a kernel feature space, commonly named a rotation, rather than performing a dimensionality reduction. Because new feature axes and KFDA projections are parallel, decision trees are used as a base classifier. To compare the proposed method with existing ensemble methods, we apply these to real datasets from the UCI and KEEL repositories.

## 1. INTRODUCTION

Classification is one of the most important challenges in data mining. It involves predicting the class label of a data instance based on input data. The classification has been used in numerous fields such as data mining, machine learning, pattern recognition, and artificial intelligence (Kim *et al*., 2014). Common methods for classification include decision tree, logistic regression, neural networks, and support vector machine.

An ensemble classifier is a combination of multiple classifiers whose decisions are merged to classify new data (Tsai *et al*., 2009). In this case, predictions from a single classifier's decision are combined into one final class label using (weighted) majority voting or distribution summation. In general, an ensemble classifier is superior to a single classifier because it utilizes the combined strength of single classifiers (Kang *et al*., 2015). Although all single classifiers are weak learners, where their classification performance is slightly greater than 50%, that is, the classification error of a single classifier $\varepsilon < 50\%$, they perform well in combination because of the diversity of an ensemble (Dietterich, 2000). When a single classifier is trained using a variety of samples, it appears to exhibit no significant difference in terms of its classification performance. However, by varying the training samples in an ensemble classifier, every single classifier operates differently, and the overall performance of the ensemble classifier improves.

Two conventional methods exist for constructing an ensemble classifier: bagging and boosting. Bagging uses bootstrap data with replacement. Each classifier, though trained on different data, is aggregated by majority voting. The accuracy of bagging improves as a result of two factors. First, the distribution of the bootstrap data approximates that of the original data. The other factor is the unstable procedure: a small change in bootstrap data causes a significant change in a single classifier, i.e., diversity (Breiman, 1996). To reinforce and utilize diversity, a variant of decision tree-based bagging known as the random forest is proposed. Random forest, which is an ensemble method for classification and regression, does not use all available features; rather it randomly creates a classifier using a decision tree after the feature subset is configured. Bootstrap data and random features yield improved classification performance (Breiman, 2001). In the R package, the number of features $p$ is determined by the square root of $p$ as a default. This method is known to exhibit excellent performance with a majority of data (Ahn *et al*., 2007).

The basic idea of boosting is to combine rough inaccurate rules of thumb. Determining many rough rules of thumb is easier than identifying a highly accurate single classifier. A representative ensemble method of boosting is AdaBoost (Friedman *et al*., 2000), which is a method that improves performance by changing the distribution of the training data during the formation of the weighted majority voting over ensembles. Using this method, we define a weak learner as a method for determining rules of thumb. When applying a weak learner to train the data, we assign more weight to data that is misclassified and lower weight to data that has been well-classified by its former learner. That is, new classifiers focus on difficult cases. Hence, the same process repeats unless a certain threshold is achieved, such that a weak classifier becomes a strong classifier (Freund and Schapire, 1995, 1996).

These methods, however, have faced the accuracy-diversity dilemma problem. AdaBoost generally exhibits superior performance compared to bagging and has a performance comparable to the random forest; however, it has limitations in raising the accuracy-diversity at the same time. Random forest, which is generally accepted as having the best performance, has a similar problem. Hence, Rotation Forest was proposed to solve the dilemma by giving up the explanatory power of the random forest.

Rotation Forest, which is a variant of random forest and is known for superior classification performance compared to general tree-based ensemble classifiers, uses principal component analysis (PCA) as a feature extraction method (Rodriguez *et al*., 2006). To create an ensemble that promotes diversity, bootstrap is used, and variable sets are randomly divided into $K$ subsets. We apply a bootstrap to the divided $K$ subsets and perform a rotation using PCA. The rotation matrix can be obtained by rearranging the rotated original data. All results derived from the tree are aggregated through majority voting. However, when the data is multiclass because PCA does not utilize the class information, there may be performance degradation issues when creating tree-based ensembles. Therefore, superior approaches have been proposed (Chen *et al*., 2014). There are variations of tree-based ensembles using kernel PCA (KPCA) or linear discriminant analysis (LDA). In the former case, KPCA was used as a feature extraction method (Shim *et al*., 2014). However, as mentioned earlier, KPCA does not consider class information of the data. In the latter case, an LDA, despite considering class information, can exhibit reduced performance when the structure of data is complicated, or the distinguishing information is in the variance, not the center, of the data (Chen *et al*., 2014).

In this study, we propose a kernel fisher discriminant analysis-based ensemble classifier called KFDA Forest, which aims at building an improved diverse ensemble classifier. We use KFDA as a feature extraction method. The proposed method is explained in Section 2. In Section 3, we present an experimental design and compare the results of the proposed method with those of current tree-based ensemble methods. Our comparison includes an assessment of bagging, AdaBoost, random forest, Rotation Forest and Canonical Linear Discriminant Analysis (CLDA) Forest. In Section 4, we discuss our experimental results in detail, provide a conclusion, and suggest directions for future research.

## 2. TREE-BASED ENSEMBLE CLASSIFIER USING KFDA

In this section, we propose a feature extraction method and KFDA-based tree ensemble classifier called KFDA Forest, in contrast to the current rotation forest. We describe the reason for the proposed method based on the literature and simulation study. We illustrate the proposed method in detail.

### 2.1 KFDA-based Feature Extraction

The main difference between a Rotation Forest and KFDA Forest is the feature extraction method, i.e., the rotation method. Whereas PCA is used as a rotation method in Rotation Forest, KFDA is used to map the original data into a higher dimensional feature space in a KFDA Forest.

PCA is an unsupervised learning method that determines a new orthogonal basis while preserving the variance of the original data to the maximum possible. Each new basis, which is generated by a linear combination of the original data's feature, is mutually orthogonal such that we can consider a new basis as a span feature. Therefore, PCA can be used as a dimensionality reduction method. However, this feature extraction method can be impractical because it does not use the class information of the input data. It is unlikely that the result of maximizing the variance through PCA necessarily leads to acceptable classification performance (Scholkopft and Mullert, 1999). The best discriminant cannot be guaranteed because any new basis does not consider the class information (Yeung *et al*., 2006).

Conversely, LDA uses the class information of the input data to determine the best discriminant vectors with a one-dimensional projection, which maximizes the between-class variance while minimizing the within-class variance (Fisher, 1936; Friedman *et al*., 2001; Yang *et al*., 2004). When input data are projected onto the discriminant vector, data are more likely to be separable than the original input space. Although LDA assumes that data are normally distributed, applying LDA, even if the input data does not satisfy the assumption of normality, is possible (Friedman *et al*., 2001).

In this study, we use KFDA, which is a generalized version of LDA, as a feature extraction method (Mika, 2003). This is because data can be effectively classified according to the class information, even with a non-linear data structure. Furthermore, KFDA has been proven to demonstrate superior performance compared to LDA in diverse areas and is the same as a combination of KPCA and LDA (Yang *et al*., 2004). Although KFDA can be used as a classification method, KFDA is not used for classification in our study and is used only as a feature extraction method for the rotation.

The following explanation is based on the description of Mika's method (Mika, 2003). Suppose that we are given data $X_1 = \{x_1^1, x_2^1, \ldots, x_{l_1}^1\}$ and $X_2 = \{x_1^2, x_2^2, \ldots, x_{l_2}^2\}$, which are samples from two classes. Let $\Phi$ be a kernel mapping to kernel feature space $F_k$. To determine the linear discriminant in $F_k$, we must maximize the following:

$$J(w) = \frac{w^T S_B^{\Phi} w}{w^T S_W^{\Phi} w} \quad (1)$$

where $w \in F_k$ and $S_B^{\Phi}$ and $S_W^{\Phi}$ are the corresponding matrices in $F_k$, that is:

$$S_B^{\Phi} = (m_1^{\Phi} - m_2^{\Phi})(m_1^{\Phi} - m_2^{\Phi})^{\mathrm{T}} \quad (2)$$

$$S_W^{\Phi} = \sum_{i=1,2} \sum_{x \in X_i} (\Phi(x) - m_i^{\Phi})(\Phi(x) - m_i^{\Phi})^{\mathrm{T}} \quad (3)$$

with $m_i = \frac{1}{l_i}\sum_{j=1}^{l_i} \Phi(x_j^i)$. To solve this problem, we use the kernel trick and the projection of *x* onto *w* given by:

$$w \cdot \Phi(x) = \sum_{i=1}^{l} \alpha_i k(x_i \cdot x) \quad (4)$$

The kernels $k(x_i \cdot x)$ compute a dot-product in some feature space $F_k$. The projection is regarded as a set of features and the projection coefficient is used to train and test the data. To summarize KFDA: KPCA is employed to map the input space to the kernel feature space and LDA is used as a feature extraction method for deriving the linear discriminants (LDs).

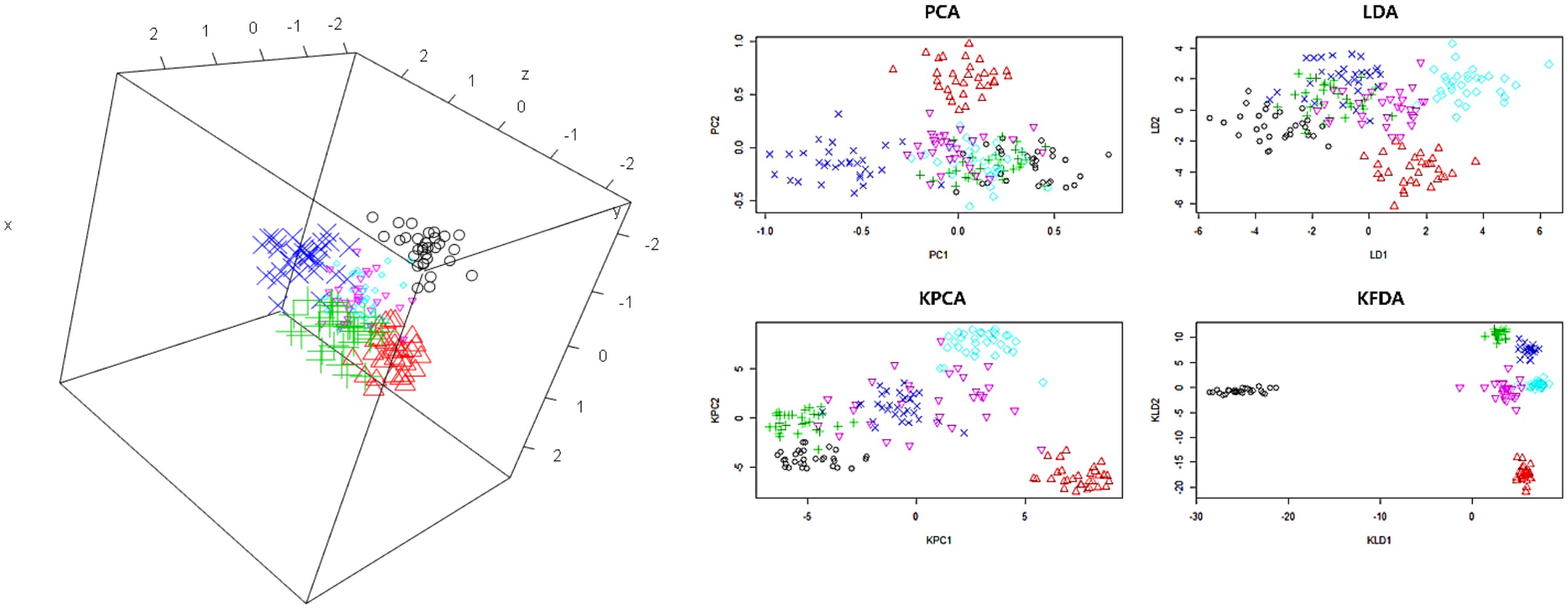

Figure 1. Distribution of 4D simplex data from the R package called *mlbench* (Leisch and Dimitriadou, 2010), in PCA, LDA, KPCA, and KFDA spaces. For clarity, only patterns from the first two subjects are plotted

The left side of Figure 1 illustrates the 4D simplex simulation data in the *mlbench* package. The right side of Figure 1 displays the results of PCA, KPCA, LDA, and KFDA for the 4D simplex simulation data. Figure 1 shows similar results to Li *et al.* (2003). In other words, LDA is superior to PCA and KPCA but not to KFDA. KFDA offers the best separation performance among the four methods. PCA and LDA are non-separable and not significantly different. KPCA is relatively more separable, and KFDA is considerably more separable than KPCA because it utilizes the class information. Therefore, as mentioned above, it is more effective to use KFDA as a feature extraction method; thus, we use KFDA as the feature extraction method for KFDA Forest. Figure 2 presents the framework of the proposed method. After extracting the features by applying KFDA to the training data obtained through the bootstrapping process, each tree learns the optimal parameters using the validation data for the kernel. The results of the learned trees are integrated into the ensemble classifier using majority voting.

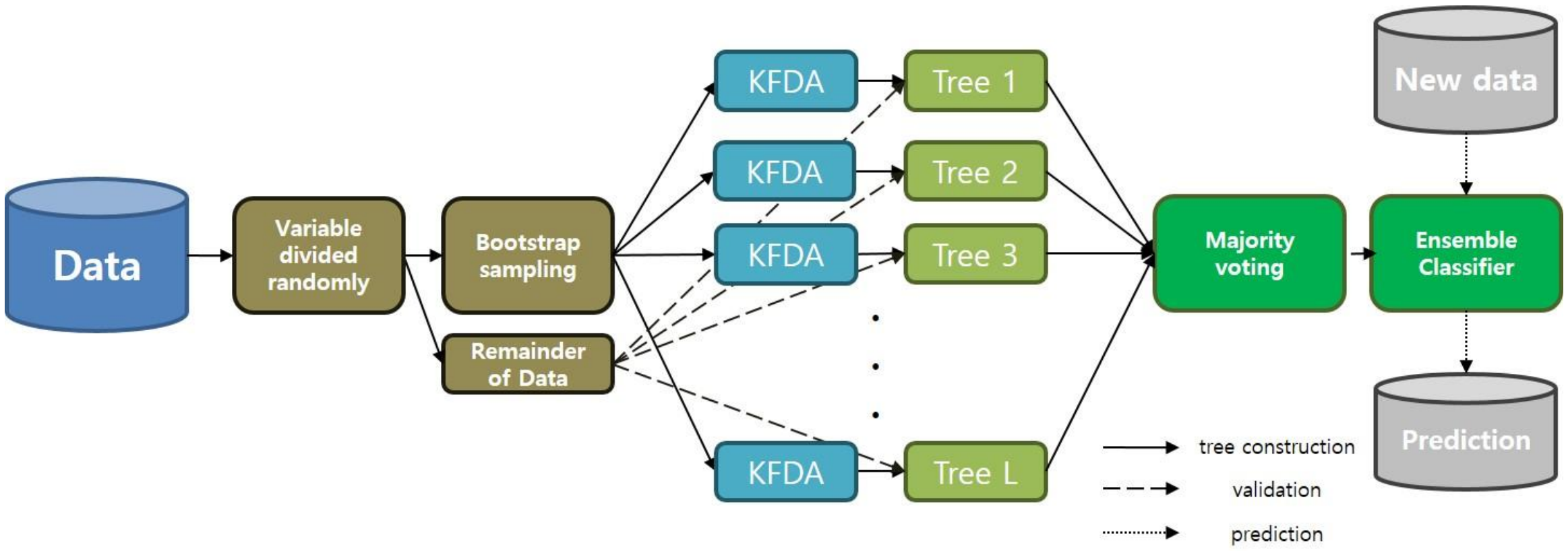

Figure 2. The framework of the proposed method

### 2.2 KFDA-based Tree Ensemble Classifier: KFDA Forest

Because the results are projected onto discriminant vectors that are aligned with the LDs (axes), using the decision tree-based ensemble method that divides the axes vertically is reasonable (Cho, 2008). Let $X = \left[x_1, x_2, \dots, x_p\right]^T$ be the training data, which is composed of $n$ instances and $p$ features. Let $Y = [y_1, y_2, \dots, y_n]^T$ be the class label vector of the training data for classification, where $y \in \{1, \dots, B\}$, with $B$ being the number of the class labels. $D_1, D_2, \dots, D_L$ are denoted by ensemble classifiers and $F$ denotes the training feature set. The method for deriving a training feature set and classifier $D_i$ is as follows.

1. $F_{i,j}$ denotes the *j*th subset of the feature training set of training data for classifier $D_i$. For each subset, draw a bootstrap resample with replacement $X_{i,j}$ from $F_{i,j}$, where the sampling size is 75% of the number of instances. This is selected for later comparison with existing methods such as Rotation Forest. The remaining training data then represent the automatic validation data for determining the optimal parameter in the kernel method.
2. Conduct KFDA on $X_{i,j}$ and obtain a coefficient matrix $C_{i,j}$ In this case, the optimal parameter for each model is conducted by a grid search.
3. Arrange the optimal coefficient matrix $C_{i,j}$ where the highest accuracy is obtained through the validation into block diagonal matrix $R_i$ as below:

$$R_i = \begin{pmatrix} C_{i,1}^{(1)}, C_{i,1}^{(2)}, \dots, C_{i,1}^{(n_1)} & 0 & \cdots \\ 0 & \ddots & \vdots \\ \vdots & \cdots & C_{i,j}^{(1)}, C_{i,j}^{(2)}, \dots, C_{i,j}^{(n_j)} \end{pmatrix} \quad (5)$$

4. Store the information of the features of each subset such that the matrix multiplication calculation can execute in a test phase. Then, construct a rearranged matrix $R_i^a$, which is a matrix with an order of original features, and arrange the coefficient matrix of $C_{i,j}$ into a single rotation matrix.
5. Build new training data for ensemble classifier $D_i$ using ($XR_i^a$, $Y$).

Because of the kernel method, in some cases, $C_{i,j}$ can have a different number of features $n_j$ for classifier $D_i$. The feature size of $C_{i,j}$ is determined by the discriminant function, which is equal to the number of variables or number of classes minus one, whichever is smaller (Tinsley *and Brown, 2000;* Poulsen and French, 2008). Therefore, if $C_{i,j}$ has a different length of features in the *j*th subset, then trim $R_i$'s dimension to the minimum number of features to rotate the original input data. When we create each tree, a new feature subset and validation set are generated to conduct KFDA. Through the validation, we determine the optimal coefficient matrix and aggregate the decisions of the trees by majority voting. Figure 3 presents the pseudocode for KFDA Forest.

***Training Phase***
**Given:**
- $X$: Training Data
- $Y$: Labels of the training data
- *L*: Number of classifiers (i.e., ensemble size)
- *K*: Number of subsets

**Output: an ensemble of decision trees** $D_1, D_2, \dots, D_L$
- For *i = 1 ,…, L*
- In training dataset $X$, randomly divide *F* (the training feature set) into *K* subsets: $F_{i,j}$ (for *j = 1 … K*)
    - Prepare the rotation matrix $R_i^a$:
    - For *j = 1 ,…, K* do
        - Make a bootstrap sample $X_{i,j}$ from $F_{i,j}$ (with a sample size of 75% of the number of instances in $X_{i,j}$).
        - Remaining data are automatically established as a validation set $X_{val}$

          ***Validation phase***
            - For sigma = *1e-7 ,…, 10* do
                - Apply KPCA plus LDA on the sample $X_{i,j}$ to obtain a coefficient matrix $C_{i,j}$
                - Fit the model of the coefficient matrix and compare the accuracy of each model using a validation set $X_{val}$
                - Obtain the best model and sigma using a grid search.
        - Arrange the best coefficient matrix of $C_{i,j}$ into a single rotation matrix
        - Store the information of the features of each subset and construct the rearranged rotation matrix $R_i^a$ by rearranging based on the rows of best $C_{i,j}$ such that they match the original order of the features in *F*
- Build a tree-based classifier $D_i$ using ($XR_i^a$, *Y*)

***Classification Phase***
**Given:** Data instance *x*, ensemble $D_1, D_2, \dots, D_L$
**Output:**
For a given data instance *x*, the predicted class label from ensemble classifier *D* is:

$$D = \frac{1}{L}\sum_{i=1}^{L} D_i(XR_i^a)$$

Figure 3. Pseudocode for KFDA Forest

## 3. EXPERIMENTS

To evaluate the performance of the proposed method, experiments using data are necessary. In this section, we conduct two types of experiments: a simulation and a real-data experiment. The simulation study was conducted to explore the possibility of the proposed method; the real-data experiment was performed to confirm the effectiveness of the proposed method when applied to real data.

### 3.1 Simulation Study

Prior to conducting a full-scale experiment, we had to ensure that the proposed methods were feasible using data with a non-linear data structure. In this study, we compared the results of the proposed method with those of the current typical tree-based ensemble methods of bagging, AdaBoost, random forest, Rotation Forest, and CLDA Forest. An unpruned decision tree was employed for all base classifiers because it yields improved results (Rodriguez *et al*., 2006). R packages called *adabag* and *randomForest* were used to implement bagging, AdaBoost, and random forest. A decision tree function called *rpart* in the R package, which is used in classification and regression tree methods was implemented for Rotation Forest, CLDA Forest, and KFDA Forest.

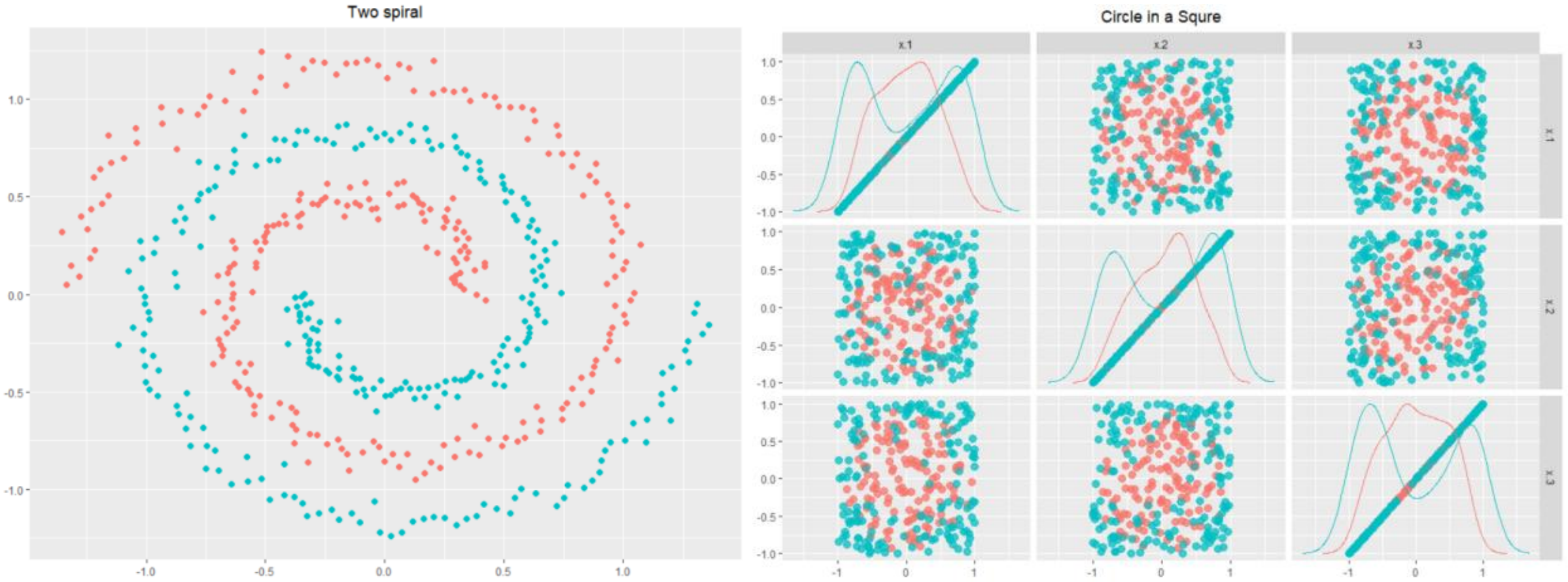

Figure 4. Two-spiral and circle-in-a-square data plot

We implemented KFDA using the most popular kernel, the radial basis function (RBF) and determined the optimal sigma using a grid search. The simulated data from the R package called *mlbench* was used to validate results and is illustrated in Figure 4. A description of the simulated data is provided in Table 1.

Table 1. Description of Two-spiral and circle-in-a-square data

| **Dataset** | **Instances** | **Dimensions** | **Classes** | **Source** |
|---|---|---|---|---|
| Two spiral | 500 | 2 | 2 | *mlbench* |
| Circle in a Square | 500 | 3 | 2 | *mlbench* |

As indicated in Figure 4, these two data sets are representative simulated data of the *mlbench* package; the structure of the data is complicated and has a binary class. Five iterations of five-fold cross-validation were conducted to compare the performance. The number of subsets $K$ was fixed to one because the dimension of the data was overly small. The ensemble size was arbitrarily set to 50. We then calculated the accuracy and standard error. The simulation results are displayed in Table 2.

Table 2. Results of simulated data based on the 5 × five-fold cross-validation with ensemble size = 50

| **Dataset** | **Measurement** | **Method** | | | | | |
|---|---|---|---|---|---|---|---|
| | | **Bagging** | **AdaBoost** | **Random Forest** | **Rotation Forest** | **CLDA Forest** | **KFDA Forest** |
| Two spiral | Accuracy (Standard error) | 0.9676 (0.0068) | 0.9708 (0.0041) | 0.9712 (0.0074) | 0.9244 (0.1602) | 0.9016 (0.0142) | **0.9984* (0.0008)** |
| Circle in a Square | Accuracy (Standard error) | 0.8913 (0.0112) | 0.9167 (0.0108) | 0.8780 (0.0117) | 0.8287 (0.0084) | 0.8727 (0.0192) | **0.9433* (0.0058)** |

*Bold denotes the highest accuracy

The simulation results demonstrate that the proposed method outperformed all other compared methods. AdaBoost and random forest demonstrated acceptable performance; Rotation Forest had the worst results, which suggests that applying PCA in some cases may result in poor results in classification problems. For the proposed approach, the standard error and accuracy for each dataset were the best among all classifiers. In this regard, we can conclude that the proposed method can outperform the others when the data structure is non-linear.

### 3.2 Experiment with Real Data

An experiment was performed on 21 real datasets. Table 3 describes the datasets used in the comparison. All these datasets were from the UCI and KEEL repositories (Blake and Merz, 1998; Alcalá *et al*., 2010). To analyze the proposed method in detail, we classified the types of variables and composed data with different instances and dimensions. The ratio of the binary and multi-class datasets was almost 1:1 because we wanted to compare the performance of the proposed method with that of the other methods according to the class type.

Table 3. Twenty-one real datasets

| Dataset | Instances | Dimensions | Discrete features | Continuous features | Classes | Source |
|---|---|---|---|---|---|---|
| Australian | 690 | 14 | 9 | 5 | 2 | KEEL |
| Balance | 625 | 4 | 0 | 4 | 3 | UCI |
| Bands | 365 | 19 | 2 | 17 | 2 | KEEL |
| Bupa | 345 | 6 | 0 | 6 | 2 | KEEL |
| Flare | 1066 | 11 | 11 | 0 | 3 | KEEL |
| Glass | 214 | 9 | 0 | 9 | 6 | KEEL |
| Haberman | 305 | 3 | 0 | 3 | 2 | KEEL |
| Heart-statlog | 270 | 13 | 0 | 13 | 2 | UCI |
| Ionosphere | 351 | 34 | 0 | 34 | 2 | UCI |
| Iris | 150 | 4 | 0 | 4 | 3 | UCI |
| Movement_libras | 360 | 90 | 0 | 90 | 15 | KEEL |
| New thyroid | 215 | 5 | 0 | 5 | 3 | KEEL |
| Pima Indians Diabetes | 768 | 8 | 0 | 8 | 2 | UCI |
| Post-operative | 86 | 7 | 7 | 0 | 2 | KEEL |
| Segment | 2310 | 18 | 0 | 19 | 7 | UCI |
| Sonar | 208 | 60 | 0 | 60 | 2 | UCI |
| Spectf heart | 267 | 44 | 0 | 44 | 2 | UCI |
| Vehicle | 846 | 18 | 0 | 18 | 4 | UCI |
| Vowel | 990 | 10 | 0 | 10 | 11 | KEEL |
| Wdbc | 569 | 30 | 0 | 30 | 2 | KEEL |
| Wine | 176 | 13 | 0 | 13 | 3 | UCI |

In the real dataset experiments, we set the ensemble size to 50 and conducted ten iterations of five-fold cross-validation to compare the accuracy of all the classifiers. The parameters of bagging, AdaBoost, and random forest were maintained at the default settings except for the unpruned tree setting. In Rotation Forest and KFDA Forest, the optimal number of subsets $K$ was chosen from one to four using validation. The ensemble size $L$, which is regarded as a hyper-parameter, was arbitrarily fixed to 50 for all classifiers. The performance measure of the ensemble classifiers was average accuracy and standard deviation. We calculated the standard deviation based on 10-iteration groups. The experimental results are presented in Table 4, and a comparison of the results by class type is displayed in Table 5.

Table 4. Results of classification accuracy and standard deviation for 21 real datasets

| Dataset | Method | | | | | |
|---|---|---|---|---|---|---|
| | Bagging | AdaBoost | Random Forest | Rotation Forest | CLDA Forest | KFDA Forest |
| Australian | 0.8802 (0.0065) | **0.8949 (0.0076)*** | 0.8931 (0.0088) | 0.8704 (0.0062) | 0.8768(0.0036) | 0.7940 (0.0141) |

| Balance | 0.8468 (0.0067) | 0.8727 (0.0093) | 0.8789 (0.0073) | 0.8800 (0.0087) | 0.9123(0.0047) | **0.9210 (0.0066)*** |
|---|---|---|---|---|---|---|
| Bands | 0.8087 (0.0086) | 0.8325 (0.0182) | **0.8376 (0.0147)*** | 0.8191 (0.0132) | 0.7203(0.0108) | 0.7874 (0.0197) |
| Bupa | 0.7742 (0.0127) | 0.7858 (0.0117) | **0.7954 (0.0201)*** | 0.7715 (0.0092) | 0.7131(0.0124) | 0.7328 (0.0169) |
| Flare | 0.7490 (0.0048) | 0.7433 (0.0067) | 0.7467 (0.0051) | 0.3300 (0.0001) | 0.7422(0.0060) | **0.7494 (0.0004)*** |
| Glass | 0.7436 (0.0162) | **0.7785 (0.0216)*** | 0.7677 (0.0120) | 0.6981 (0.0166) | 0.6200(0.0167) | 0.6944 (0.0147) |
| Haberman | 0.7542 (0.0145) | 0.7498 (0.0250) | 0.7658 (0.0142) | 0.7381 (0.0123) | 0.7521(0.0067) | **0.7813 (0.0129)*** |
| Heart-statlog | 0.8195 (0.0128) | 0.7997 (0.0153) | 0.8282 (0.0116) | 0.8147 (0.0142) | 0.8354(0.0076) | **0.8405 (0.0077)*** |
| Ionosphere | 0.9173 (0.0088) | 0.9379 (0.0055) | 0.9351 (0.0054) | 0.9363 (0.0064) | 0.8732(0.0085) | **0.9394 (0.0052)*** |
| Iris | 0.9444 (0.0085) | 0.9476 (0.0057) | 0.9499 (0.0059) | 0.9667 (0.0057) | 0.9751(0.0063) | **0.9794 (0.0065)*** |
| Movement_libras | 0.7500 (0.0168) | 0.8169 (0.0151) | 0.8383 (0.0122) | 0.7980 (0.0050) | 0.6963(0.0232) | **0.8661 (0.0157)*** |
| New thyroid | 0.9590 (0.0120) | 0.9595 (0.0113) | **0.9637 (0.0065)*** | 0.9488 (0.0090) | 0.9466(0.0063) | 0.9567 (0.0073) |
| Pima Indians Diabetes | 0.7668 (0.0111) | 0.7392 (0.0103) | 0.7585 (0.0054) | 0.7479 (0.0076) | **0.7851(0.0063)*** | 0.7634 (0.0064) |
| Post-operative | 0.6042 (0.0245) | 0.6206 (0.0237) | 0.6687 (0.0137) | **0.7395 (0.0004)*** | 0.6940(0.0263) | 0.6790 (0.0090) |
| Segment | 0.9242 (0.0028) | **0.9843 (0.0012)*** | 0.9779 (0.0024) | 0.9310 (0.0051) | 0.9248(0.0032) | 0.9625 (0.0020) |
| Sonar | 0.7836 (0.0197) | 0.8073 (0.0283) | 0.8191 (0.0111) | 0.8352 (0.0220) | 0.7505(0.0213) | **0.8510 (0.0160)*** |
| Spectf heart | 0.8195 (0.0128) | 0.8041 (0.0149) | 0.8027 (0.0121) | **0.8229 (0.0131)*** | 0.7523(0.0124) | 0.7942 (0.0003) |
| Vehicle | 0.6805 (0.0082) | 0.7411 (0.0058) | 0.7324 (0.0068) | 0.7263 (0.0068) | 0.6732(0.0078) | **0.7726 (0.0129)*** |
| Vowel | 0.6787 (0.0051) | 0.8963 (0.0073) | 0.9494 (0.0081) | 0.7704 (0.0090) | 0.6913(0.0082) | **0.9773 (0.0035)*** |
| Wdbc | 0.9573 (0.0024) | **0.9738 (0.0023)*** | 0.9653 (0.0050) | 0.9622 (0.0058) | 0.9736(0.0045) | 0.9560 (0.0037) |
| Wine | 0.9630 (0.0081) | 0.9659 (0.0093) | 0.9774 (0.0059) | 0.9591 (0.0094) | **0.9806(0.0123)*** | 0.9801 (0.0063)* |

Number in parentheses denotes the standard deviation of ten iterations.
*bold denotes the highest accuracy

Among all datasets, KFDA was superior 10 of 21 times, AdaBoost and random forest were superior four and three out of 21 times, respectively, and the remaining methods were superior four out of 21 times. In the case of Rotation Forest and KFDA Forest, the best performance of each method based on the optimum value of K is displayed in Table 4 (*K* was varied from one to four). When *K* was greater than or equal to two, features were selected randomly, and the ensemble diversity also increased in these methods, thus yielding better performance.

Table 5. Comparison of results according to class type

| **Class type** | **# of datasets** | **Method** | | | | | |
|---|---|---|---|---|---|---|---|
| | | **Bagging** | **AdaBoost** | **Random Forest** | **Rotation Forest** | **CLDA Forest** | **KFDA Forest** |
| Binary class | 11 | 0 | 2 | 2 | 2 | 1 | **4** |
| Multi-class | 10 | 0 | 2 | 1 | 0 | 1 | **6** |
| Total number of wins | | 0 | 4 | 3 | 2 | 2 | **10** |

Number in bold denotes the highest accuracy

In the case of binary classes, the proposed method was superior four out of 11 times. Specifically, AdaBoost, random forest, and Rotation Forest outperformed bagging and CLDA Forest but performed worse than the proposed method. Bagging usually performs poorly on binary datasets. The experimental results for the binary class using KFDA Forest were a consequence of the characteristics of LDA. The discriminant function of the binary class data was always one, regardless of the dimensions of the original input data. Therefore, it is likely that binary data was not always separable when applying KFDA Forest.

However, the proposed method outperformed the other methods with respect to the multi-class datasets. In this case, the proposed method was superior to six out of 10 times. Because of the limitations of CLDA, CLDA Forest also showed lower performance than the proposed method. The experimental results of the multi-class classifications for KFDA Forest were based on the number of classes. The greater the number of classes present, the greater the dimensions of the subsets and the discriminant function, as mentioned previously. For this reason, using KFDA, we should utilize as many features (LDs) as possible so that KFDA performance improves significantly. The experimental results from the other methods on multi-class datasets were similar to their binary class performances.

Table 6. Comparison of results according to variable type

| Variable type | # of datasets | Method | | | | | |
|---|---|---|---|---|---|---|---|
| | | Bagging | AdaBoost | Random Forest | Rotation Forest | CLDA Forest | KFDA Forest |
| Discrete variables | 2 | 0 | 0 | 0 | **1** | 0 | **1** |
| Mixed variable | 2 | 0 | **1** | **1** | 0 | 0 | 0 |
| Continuous variable | 17 | 0 | 3 | 2 | 1 | 2 | **9** |
| Total number of wins | | 0 | 4 | 3 | 2 | 2 | **10** |

Number in bold denotes the highest accuracy for the variable type.

Table 6 summarizes the number of methods that achieved the highest accuracy based on variable types. When the data variables were continuous, the KFDA Forest achieved acceptable performance; however, when the variables were discrete or nominal, the performance was not practical. It would appear that the conversion of the variables, called encoding, weakens the mapping effect through the kernel discriminant analysis for feature extraction.

### 3.3 Discussions

Table 2 shows the simulation results and Table 4 shows the results of verifying the proposed method by applying them to actual data. In addition, Tables 5 and 6 clearly show the strengths (and weaknesses) of the proposed method by summarizing the performance results based on the class type and variable type, respectively.

The fact that the various experiments reveal is that the performance of the tree ensemble method can change dramatically depending on the method used to extract features. In addition, these characteristics are highlighted in Tables 5 and 6, and there is a strong correlation between the characteristics and the results of KFDA. The significance of this experiment is that a feature extraction method that can reflect the class label information of the data is effective, and as a characteristic of the general kernel trick, the data is more likely to be separable as it increases, in general. Although it is not easy to find the optimal kernel parameters for each tree in the proposed method, the method exhibits good performance if the kernel parameters are determined through validation. This implies that there may be a specific optimum kernel for the dataset as well as the kernel parameters, and selecting (or estimating) a near-optimum set of kernel parameters may offer the better performance of the proposed method.

## 4. CONCLUSIONS

In this study, we employed KFDA and tree-based ensembles as feature extraction and classification method for a variety of real datasets. KFDA used a rotation method rather than PCA because PCA does not necessarily lead to acceptable performance for different types of data classes. As mentioned earlier, KPCA does not consider class information of the data. LDA, despite considering class information, has reduced performance when the structure of the data is complicated or the distinguishing information is in the variance, not in the center of the data. Conversely, KFDA considers both class information and a complex data structure; hence, classification performance can improve even though the data structure is non-separable in the input data space. Based on the characteristics of KFDA, the projection of data is aligned with the axes. Therefore we used a decision tree as the base classifier. By adopting the kernel method and applying it to the decision trees, different kernel feature spaces are created, which increases the ensemble diversity This allowed the development of the proposed ensemble classifier.

We compared the feasibility and performance of the proposed method through simulation and real-data experiments. On average, we confirmed that the proposed method outperformed the conventional tree-based ensemble classifiers: bagging, AdaBoost, random forest, Rotation Forest, and CLDA Forest. In particular, in the multi-class cases, the proposed method outperformed the majority of the other methods in terms of average classification accuracy. Further, it demonstrated a standard deviation less than that of the other methods in many cases.

Although the results of the experiment are notable, some limitations exist in the proposed method. First, we used only the RBF kernel to implement KFDA Forest. Hence, we were required to determine the exact nature of the performance of the proposed method compared to methods using other kernel functions. Second, comparing performance based on various ensemble sizes was necessary. If a smaller ensemble performed better, it was regarded as better. Finally, the computational time of KFDA Forest is relatively high when compared to the other tree-based ensemble classifiers. This is because of the method used for determining the optimal kernel parameter to train the ensemble classifier. For every single classifier *D,* the optimal value of the kernel parameter, sigma, was determined using a grid search. Therefore, developing a method to determine the optimal parameters efficiently to reduce the training time should be implemented.

**ACKNOWLEDGMENTS**

This work was supported by the National Research Foundation of Korea (NRF) grant funded by the Korea government (MSIT) (No. NRF-2016R1A2B4013678). This work was also supported by the BK21 Plus (Big Data in Manufacturing and Logistics Systems, Korea University) and the Samsung Electronics Co., Ltd.